\documentclass[runningheads]{llncs}

\usepackage{eccv}

\usepackage{eccvabbrv}

\usepackage{graphicx}
\usepackage{booktabs}
\usepackage{multirow}
\usepackage{multicol}
\usepackage{algorithm}
\usepackage{algorithmic}
\usepackage{appendix}
\usepackage{tabularx}
\usepackage{makecell}
\usepackage{float}
\usepackage{tabularx}
\usepackage{hyperref}

\newcommand{\X}{\boldsymbol{X}}
\newcommand{\Y}{\boldsymbol{Y}}
\newcommand{\x}{\boldsymbol{x}}
\newcommand{\y}{\boldsymbol{y}}
\newcommand{\s}{\boldsymbol{s}}

\newcommand{\W}{\boldsymbol{W}}
\newcommand{\U}{\boldsymbol{U}}
\newcommand{\Q}{\boldsymbol{Q}}
\newcommand{\q}{\boldsymbol{q}}
\newcommand{\w}{\boldsymbol{w}}

\newcommand{\R}{\mathbb{R}}

\usepackage[accsupp]{axessibility}  

\usepackage[pagebackref,breaklinks,colorlinks,citecolor=eccvblue]{}

\usepackage{orcidlink}

\begin{document}

\title{COMQ: A Backpropagation-Free Algorithm for Post-Training Quantization} 

\titlerunning{COMQ: A Backpropagation-Free Algorithm for Post-Training Quantization}


\author{Aozhong Zhang\inst{1} \and
Zi Yang\inst{1} \and
Naigang Wang\inst{2} \and
Yingyong Qi\inst{3} \and 
Jack Xin\inst{3} \and 
Xin Li\inst{4} \and 
Penghang Yin\inst{1}}

\authorrunning{Aozhong et al.}

\institute{Department of Mathematics and Statistics, University at Albany, SUNY \\
\email{\{azhang3, zyang8, pyin\}@albany.edu} \and
IBM T. J. Watson Research Center \\
\email{nwang@us.ibm.com} \and
Department of Mathematics, University of California at Irvine\\
\email{\{yqi, jack.xin\}@uci.edu} \and
Department of Computer Science, University at Albany, SUNY \\ 
\email{xli48@albany.edu}}

\maketitle

\begin{abstract}
  Post-training quantization (PTQ) has emerged as a practical approach to compress large neural networks, making them highly efficient for deployment. However, effectively reducing these models to their low-bit counterparts without compromising the original accuracy remains a key challenge. In this paper, we propose an innovative PTQ algorithm termed COMQ, which sequentially conducts \underline{\bf co}ordinate-wise \underline{\bf m}inimization of the layer-wise reconstruction errors. We consider the widely used integer quantization, where every quantized weight can be decomposed into a shared floating-point scalar and an integer bit-code. Within a fixed layer, COMQ treats all the scaling factor(s) and bit-codes as the variables of the reconstruction error. Every iteration improves this error along a single coordinate while keeping all other variables constant. COMQ is easy to use and requires no hyper-parameter tuning. It instead involves only dot products and rounding operations. We update these variables in a carefully designed greedy order, significantly enhancing the accuracy. COMQ achieves remarkable results in quantizing 4-bit Vision Transformers, with a negligible loss of less than 1\% in Top-1 accuracy. In 4-bit INT quantization of convolutional neural networks, COMQ maintains near-lossless accuracy with a minimal drop of merely 0.3\% in Top-1 accuracy. The code is available at \href{https://github.com/AozhongZhang/COMQ}{https://github.com/AozhongZhang/COMQ} 
  \keywords{Post-training quantization \and Coordinate descent \and Layer-wise reconstruction}
\end{abstract}

\section{Introduction}
\label{sec:intro}

Over the past decade, deep learning has enjoyed remarkable success in a wide range of fields \cite{krizhevsky2012imagenet, faster_rcnn, silver2016mastering, jumper2021highly,devlin2018bert,radford2018improving,ho2020denoising}. Deep neural networks (DNNs) are scaled to unprecedented sizes for better performance, resulting in models with billions or even trillions of parameters. With the increasing demand to deploy these models on resources-constrained devices, substantial efforts have been dedicated to advancing model compression techniques \cite{li2020block, he2017channel, li2016pruning, rastegari2016xnor, uhlich2019mixed, zhou2016dorefa} to train lightweight DNNs without sacrificing achievable accuracy. Among these techniques, quantization has recently garnered significant attention. Quantization techniques involve representing the weights and activations of DNNs with low precision using just a few bits (e.g., 4-bit). Quantized DNNs offer a substantially reduced memory footprint and rely on fixed-point or integer arithmetic during inference, leading to accelerated and efficient deployment.

Quantization methods can be categorized into two primary categories: Quantization Aware Training (QAT) \cite{rastegari2016xnor, hubara2017quantized, wang2017fixed, yin2019blended,yin2018binaryrelax,zhou2016dorefa,krizhevsky2017imagenet, yin2019understanding} and Post-Training Quantization (PTQ) \cite{li2021brecq, nagel2020up, frantar2022optimal, lin2023bit, hubara2020improving, wei2022qdrop, wang2022deep}. In general terms, QAT is designed to globally minimize the conventional training loss of the model for quantization parameters. It involves tackling a formidable nonconvex minimization problem with a discrete nature. QAT requires an all-encompassing training pipeline and computational cost that is at least on par with regular full-precision models. In contrast, PTQ directly applies low-precision calibration to a pre-trained full-precision model. Computationally, PTQ aims to identify an optimal quantized model locally by minimizing a simplified surrogate loss, so it enjoys significantly reduced algorithmic complexity and appears to be a faster and more resource-efficient process. On the downside, PTQ suffers a heavier performance degradation than QAT, especially when it comes to low-bit quantization of Vision Transformers (ViTs), which has received much recent attention \cite{dosovitskiy2020image,touvron2021training,liu2021swin, liu2021post}.

\smallskip

\noindent {\bf Motivation.} Unlike QAT, PTQ enjoys the benefits of resource efficiency. The downside of PTQ includes potential accuracy drop, sensitivity to model and data distribution, and limited flexibility in precision levels. Therefore, it is desirable to pursue some middle ground between QAT and PTQ by preserving the cost benefits of PTQ without sacrificing the accuracy. Optimization theory offers a rich weaponry for striking an improved tradeoff between cost and performance in many vision systems. For model compression, a greedy path-following mechanism was developed for PTQ of neural networks with provable guarantees in \cite{zhang2023post}; a novel bit-split optimization approach was developed in \cite{wang2022optimization} to achieve minimal accuracy degradation based on the analysis of the quantization loss in the final output layer. Inspired by these recent advances, we advocate an optimization-based approach to PTQ based on coordinate descent \cite{nesterov2012efficiency,wright2015coordinate,hsieh2008dual} that minimizes the objective functions along the coordinate directions iteratively. 

\smallskip

\begin{figure*}[htbp]
\centering 
\includegraphics[width=0.68\textwidth]{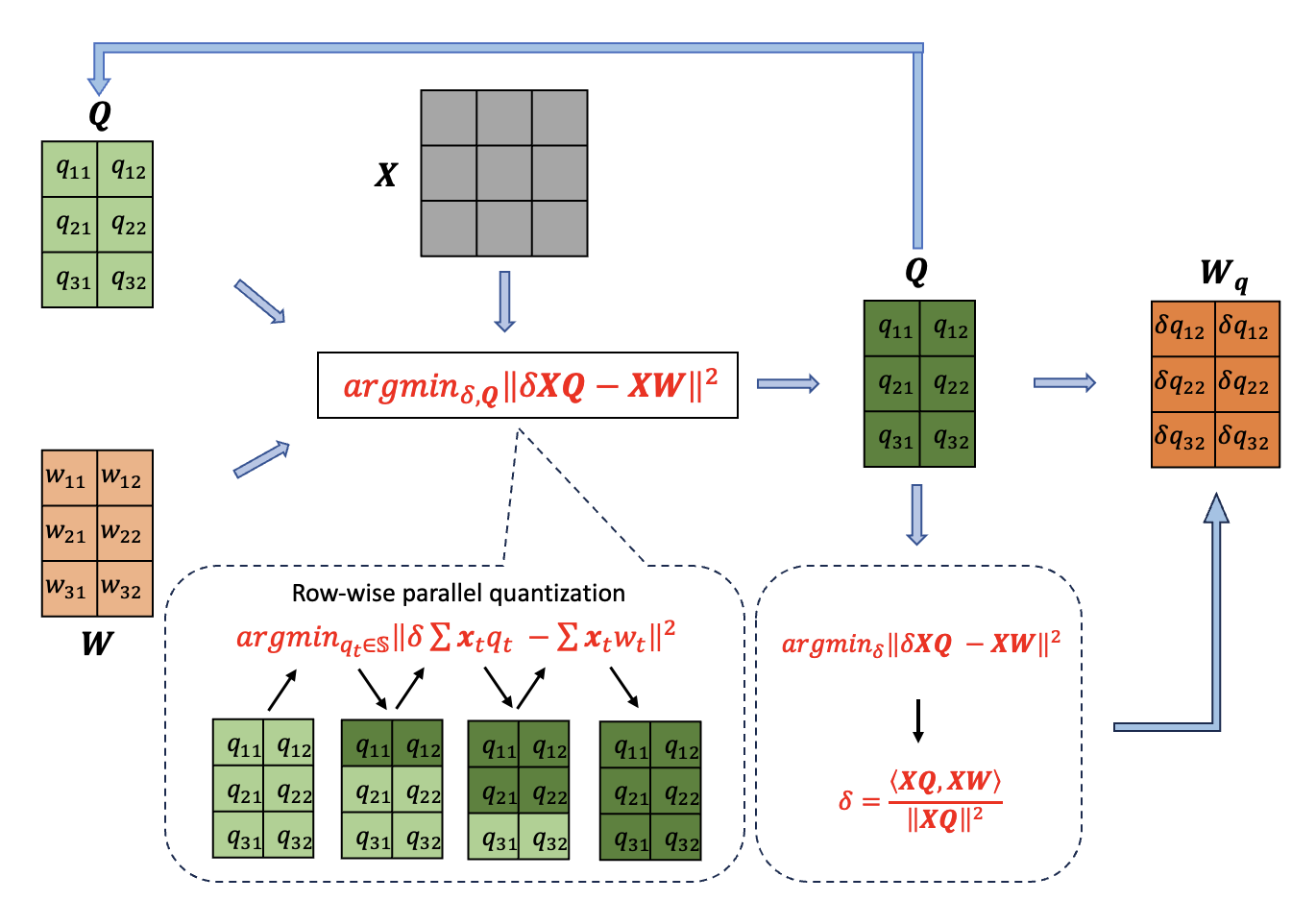} 
\caption{The workflow of COMQ for per-layer quantization.} 
\vspace{-0.2in}
\label{Fig1} 
\end{figure*}

\noindent {\bf Contribution.} In this work, we introduce COMQ, a post-training quantization method that performs uniform weight quantization on a layer-by-layer basis (see Figure \ref{Fig1}). Similarly to existing works \cite{nagel2020up,hubara2020improving, frantar2022optimal,zhang2023post}, our main objective is to minimize the layerwise squared error $\| \X \W_q - \X\W\|^2$ with respect to quantized weights $\W_q$. To efficiently address optimization, COMQ enforces the decomposition $\W_q = \delta \cdot \Q$ with $\delta$ being the full-precision scalar(s) and $\Q$ storing the integer bit-codes, and it then minimizes this error over the new variables through a coordinate-wise minimization procedure. Adhering to a greedy selection rule, we select one variable at a time, whether scaling factor or bit-code, to update while maintaining the others at their most recent states.  Unlike recent works \cite{frantar2022optimal,li2021brecq,nagel2020up} that require back-propagation or estimation of the Hessian inverse to minimize the reconstruction error, COMQ solves a sequence of minimization of \emph{univariate quadratic functions} which enjoy closed-form minimizers. This leads to backpropagation-free iterations that primarily involve dot products and rounding operations without using any hyperparameters. 

We detail the implementation of COMQ for per-layer and per-channel weight quantization, respectively.
Our empirical results demonstrate that the proposed selection order of variables can enhance the performance at extremely low bit-widths, outperforming the standard index-based update order (the so-called cyclic order). Our experiments show that the proposed COMQ method achieves state-of-the-art performance on convolutional neural networks and Vision Transformers. Specifically, our 4-bit CNN almost reaches the accuracy of full-precision models with less than $0.05\%$ accuracy loss, and 4-bit ViT reaches less than $1\%$ accuracy loss. Moreover, our approach outperforms the existing state-of-the-art PTQ methods on CNNs with per-layer quantization and ViTs with per-channel quantization. Although our main focus is on weight quantization, we remark that the proposed framework exhibits versatility, allowing seamless extension to full quantization tasks by incorporating existing activation quantization techniques, especially that for quantizing ViTs such as \cite{li2023repq}. 

\smallskip

\noindent \textbf{Notations.} Throughout this paper, we denote vectors with bold small letters and matrices with bold capital ones.
For any two vectors $\x, \y \in \mathbb{R}^{n}$, $\langle \x, \y\rangle := \x^\top\y = \sum_{i=1}^n x_{i}y_{i}$ is their inner product. We denote by $\|\x\| := \sqrt{\langle \x, \x\rangle}$ the Euclidean norm and denote by $\|\x\|_\infty=\max_{i=1,\ldots,n} |x_i|$ the $l_\infty$ norm. Similarly for two matrices $\X, \Y \in\mathbb{R}^{m\times n}$, the inner product is given by $\langle \X, \Y\rangle := \sum_{i=1}^m \sum_{j=1}^n X_{i,j} Y_{i,j}$, and $\|\X\| := \sqrt{\langle \X, \X\rangle}$ is the Frobenius norm. Moreover, we denote the outer product of two vectors by $\x \otimes \y: = \x \y^\top\in\mathbb{R}^{n\times n}$. $\x\odot\y := (x_1 y_1, \dots, x_n y_n)\in\R^n$ denotes the Hadamard (or element-wise) product, and $\x\oslash\y := (x_1/y_1,  \dots, x_n/y_n)\in\R^n$ denotes the Hadamard division. Finally, for any positive integer $n$, $[n]:=\{1,\dots,n\}$ denotes the set of integers up to $n$.

\section{Related Works}

\subsection{Post-training quantization} 
The earlier PTQ work, DFQ \cite{nagel2019data}, is a data-free approach that operates independently of any training data. It involves minimizing the expected error between outputs of the corresponding linear layers in both the pre-trained and quantized models over the inputs. DFQ employs pre-processing steps to appropriately re-scale the weights and biases across the layers before the quantization. DFQ relies on prior information about the mean of layer inputs, a value that can be estimated from batch normalization parameters to rectify the noises introduced during the quantization process. DFQ is proven to work well on INT8 quantization but suffers noticeable accuracy degradation at lower bit-widths. Cai et al. \cite{cai2020zeroq} introduce ZeroQ, which distills the input data distribution that matches the statistics in the model's batch normalization layers. Xu et al. \cite{xu2020generative} propose to use a generative model to construct fake data for accuracy improvement. Subsequent research efforts employ a set of calibration data and perform layer-by-layer or block-by-block quantization. The general idea is to minimize layer-wise \cite{nagel2020up, hubara2020improving, hubara2021accurate, frantar2022optimal, zhang2023post} or block-wise \cite{li2021brecq} squared error between outputs of pre-trained and quantized models with respect to the associated quantized weights. To carry out the minimization procedure, a class of PTQ methods such as AdaRound \cite{nagel2020up}, BRECQ \cite{li2021brecq}, QDrop \cite{wei2022qdrop}, adopt back-propagation to minimize the layer-wise or block-wise quantization error. AdaQuant \cite{hubara2020improving,hubara2021accurate} proposed to solve integer programming for bit-allocation while
jointly learning a scaling factor.

Some initiatives have been undertaken to develop backpropagation-free algorithms for weight quantization. OBC \cite{frantar2022optimal}, also known as OPTQ \cite{frantar2022gptq} for quantizing large language models (LLMs), adapts the Optimal Brain Surgeon framework to the setting of layer-wise PTQ. OBC quantizes one weight using an analytical formula at a time and updates all remaining weights after each step. Compared with OBC, our proposed COMQ method is simpler because it only updates one parameter in each step and keeps the others untouched. Additionally, unlike OBC, COMQ does not require inverting the Hessian matrix of the layer-wise squared error in quantizing the weight. Zhang et al. \cite{zhang2023post} introduced GPFQ to efficiently learn layer-wise bit-codes sequentially, assuming that the floating scaling factors are already learned. In its practical implementation, determining the appropriate scalars requires a trial-and-error process. 
In parallel to our research, Behdin et al. explored a cyclic coordinate descent approach, QuantEease \cite{behdin2023quantease}, for PTQ. While their work focuses on LLMs, our research targets ViTs with a different algorithmic implementation. Specifically, we propose a greedy update order and introduce learnable scaling factors, which we believe offer unique benefits in terms of performance.

Developing tailored PTQ methods for ViTs while maintaining good performance poses a significant challenge and has garnered considerable attention \cite{yuan2021ptq4vit,ding2022towards,li2023repq,lin2021fq}. For examples, PTQ4ViT \cite{yuan2021ptq4vit} proposes twin uniform quantization to cope with the unbalanced distributions of activation values and a Hessianguided metric to search for scaling factors. APQ-ViT \cite{ding2022towards} proposes a calibration scheme that perceives the overall quantization disturbance block-wise. FQ-ViT \cite{lin2021fq} introduces
Powers-of-Two scale and Log-Int-Softmax quantizers for the activation quantization. RepQ-ViT \cite{li2023repq} proposes scale reparameterization methods for post-LayerNorm and post-Softmax activations, and improves the accuracy of 4-bit PTQ of ViTs to a usable level. More recently, PTQ methods have also been applied to the emerging diffusion models \cite{he2024ptqd,shang2023post} and large language models \cite{xiao2023smoothquant,frantar2022gptq}.

\subsection{Coordinate descent}

The coordinate descent method is a simple yet effective optimization algorithm widely applied to large-scale problems \cite{nesterov2012efficiency,wright2015coordinate,hsieh2008dual}. The algorithm minimizes the objective functions along the coordinate directions iteratively. For the objective function $f(x_1,\ldots,x_d)$, the coordinate descent method starts with the initial value $(x_1^0,\ldots,x_n^0)$. In the $k$-th iteration, the method sequentially solves the following problem for $i=1,\ldots,n$
\begin{equation*}
   x_i^k := \arg\min_{x_i} f(x_1^{k},\ldots,x_{i-1}^k, x_i, x_{i+1}^{k-1},\ldots,x_n^{k-1}).
\end{equation*}
In the PTQ problem, each quantized weight is represented as $\W_q = \delta \cdot \Q$. We consider the scale $\delta$ and all elements in $\Q$ as coordinates. In one step, we update one element in $\Q$ or the scale $\delta$ while fixing all other coordinates. Each subproblem in the coordinate descent method is a univariate optimization problem and has a closed-form solution in our settings. The order of solving subproblems affects the optimization performance. Cyclic order is the most widely used, cyclically iterating through the coordinates. However, the minimization problem for quantization is non-convex and exhibits a discrete nature, challenging the optimality of cyclic order. In this work, we propose to update the parameters in a carefully designed order to achieve a reduced quantization loss compared to that of cyclic order \cite{wang2022optimization}.

\section{Proposed Method}

This section presents our proposed COMQ method for post-training quantization. Our discussions mainly focus on linear layers for simplicity. A convolutional layer can be equivalently converted to a linear layer, hence the proposed method can be also applied. 
Regarding transformers \cite{vaswani2017attention}, it is natural to conceptualize the key, query, and value components of a self-attention layer as three separate linear layers. 

We consider a linear layer with matrix weight $\W$ and matrix input $\X$. For the weight $\W$, we aim to find the quantized weights $\W_q$ that minimize the following function 
\begin{equation} \label{eq:minWi}
    \min_{\W_q \in \mathcal{W}} \; \| \X \W_q - \X \W\|^2,
\end{equation}
where $\mathcal{W}$ is an appropriate set of all feasible quantized weights. The matrix input $\X$ is the feature generated from the pre-trained model and does not depend on the quantized weights from the previous layer. We propose to solve the problem \eqref{eq:minWi} by the coordinate descent algorithm. We regard all elements in the quantized weight and the scale factor as coordinates. In each step, we only solve a univariate optimization problem with respect to only one coordinate. By iteratively computing the quantized weights and corresponding scale factors, we finally find a proper quantized weight $\W_q$ that has a small quantization error and maintains good accuracy.

In this section, we will show our COMQ method for two quantization scenarios: per-layer quantization and per-channel quantization. All quantized weights with in the same layer share a common scale factor for the per-layer quantization, while different columns of the quantized weight use different scale factors for the per-channel quantization. Throughout the paper, we consider the $b$-bit asymmetric uniform quantization \cite{choi2018pact}, which takes the bit-code set of $\mathbb{S}=\{z, z+1, \ldots,z+2^{b}-1\}$, with $z$ being the so-called zero point.

\subsection{Per-layer Quantization}
\label{sc:layer}
In this subsection, we present our COMQ method for $b$-bit per-layer quantization, where the whole quantized weight matrix $\W_q$ shares the scale factor $\delta$. In this setting, the quantized weight $\W_q$ can be decomposed as 
\[
    \W_q = \delta \cdot \Q \in \mathbb{R}^{m\times n},
\]
where $\Q\in\mathbb{S}^{m\times n}$ is the integer bit-code matrix for $\W_q$. The optimization problem \eqref{eq:minWi} reduces to 
\begin{equation} \label{eq:minW}
    \min_{\delta \in \mathbb{R}, \Q\in \mathbb{S}^{m\times n}} \; \| \delta \X \Q - \X \W\|^2.
\end{equation}
It holds that $\X\Q = (\X\q_1,\ldots,\X\q_{n})$ and $\X\W = (\X\w_1,\ldots,\X\w_{n})$, where $\q_j$ and $\w_j$ are the $j$-th column of $\Q$ and $\W$ respectively. Therefore, we can rewrite the problem \eqref{eq:minW} as 
\begin{equation} \label{eq:minq sep}
    \min_{\delta, \q_1,\ldots,\q_{n}} \sum_{j=1}^{n} \|\delta \X \q_j - \X \w_j\|^2.
\end{equation}
When fixing the scale factor $\delta$, the terms $\|\delta \X \q_j - \X \w_j\|^2$ for $j=1,\ldots,n$ are \emph{independent} of each other. Hence, we can update the columns $\q_1,\ldots,\q_{n}$ \emph{in parallel}. We apply the coordinate descent method to solve the above problem \eqref{eq:minW}. Specifically, we regard the scaling factor $\delta$ and all elements in $\Q$ as individual coordinates. In each iteration, we can update the bit-codes across all columns in the same row, i.e., row-wise update, before updating $\delta$: 
\begin{equation} \label{eq:order}
\{Q_{1,j}\}_{j\in[n]} \Rightarrow \{Q_{2,j}\}_{j\in[n]} \Rightarrow \cdots \Rightarrow \{Q_{m,j}\}_{j\in[n]} \Rightarrow \delta.
\end{equation}

\noindent \textbf{Initialization}. At the beginning of our COMQ method, the scale factor $\delta^0$ should be properly initialized to capture the range of the weight matrix $\W$. Typically, the maximum value of $|\W|$ is the default selection in most quantization problems. However, this choice overlooks the impact of outliers in the weights on the overall quantization error. In order to smooth out these outliers, we consider the average infinity norm of weights across all columns $\W$. For the $b$ bits uniform quantization, we use the initialization 
$\delta^0 = \frac{1}{2^{b-1}}\frac{\sum_{1 \leq i \leq n}\|\w_i\|_{\infty}}{n}$.
Then the matrix $\Q$ is initialized as $\Q^0 = \frac{\W}{\delta^0}$. Note that the initialization $\Q^0$ above is not an actual bit-code matrix as $\W$ is float, but it will become feasible after the 1st iteration as will be shown below.

\noindent \textbf{$\Q$-Update}. Let $\Q^{k-1},\delta^{k-1}$ be the parameters after $k-1$ iterations. Without loss of generality, we focus on updating an arbitrary column of $\Q$, denoted by $\q$, where the column index $j$ is omitted for notational simplicity. 
Suppose the coordinates $q^{k}_t$ for $1\leq t\leq i-1$, have been updated in the $k$-th iteration. Note that $\X\q = \sum_{t=1}^m q_t \x_t$, with $\x_t$ being the $t$-th column of $\X$. Coordinate descent calls for solving the following problem to obtain $q^{k}_i$: 
\begin{align} \label{eq:findq}
  q^{k}_i = & \;  \arg\min_{q_i\in\mathbb{S}} \left\|\delta^{k-1} \left(q_i\x_i +   \sum_{t=1}^{i-1} q_t^k \x_t +  \sum_{t=i+1}^{m} q_t^{k-1} \x_t \right) - \X \w \right\|^2 \notag\\
  = & \; \arg\min_{q_i\in\mathbb{S}} \left\| q_i \delta^{k-1} \x_i -\s_i^k \right\|^2,
\end{align}
where $\s_i^k = \X \w - \delta^{k-1}(\sum_{t=1}^{i-1} q_t^k \x_t + \sum_{t=i+1}^{m} q_t^{k-1} \x_t)$ is a constant vector. \eqref{eq:findq} is a quadratic minimization problem if the variable $q_t$ is in the continuous domain. The global optimizer of \eqref{eq:findq} over $\mathbb{R}$ is $\frac{\left< \delta^{k-1} \x_{i}, \s^k_i \right>}{\|\delta^{k-1}\x_{i}\|^{2}}$. Therefore, we obtain the updated coordinate $q_i^{k}$ as 
\begin{equation*} 
    q^{k}_i = \text{clip} \left ( \left \lfloor \frac{\left< \x_{i}, \s^k_i \right>}{\delta^{k-1}\|\x_{i}\|^{2}} \right \rceil, z, z+2^{b}-1 \right ). 
\end{equation*}
 
 Hereby we define the quantization residuals at the $i$-th step (for the column): 
 \begin{align*}
 \boldsymbol{u}_i^k : = \sum_{t=1}^{i-1} (w_t -\delta^{k-1} q_t^k) \x_t + \sum_{t=i}^{m} (w_t - \delta^{k-1} q_t^{k-1}) \x_t = \s_i^k - \delta^{k-1}q_i^{k-1}.
 \end{align*}
 To efficiently implement the $\Q$-update as in \eqref{eq:order}, we take advantage of vectorized operations to update $\{Q_{i,j}\}_{j\in[n]}$ across all columns for each $i$. To this end, we denote by $\w_{i,:}\in\R^n$ and $\q_{i,:}\in\R^n$ the $i$-th row of $\W$ and $\Q$, respectively, and denote by $\x_{:,i}\in\R^m$ the $i$-th column of $\X$.
 Then the vectorized update of $\{Q_{i,j}^k\}_{j\in[n]}$ or $\q_{i,:}^k$ for the $k$-th iteration of COMQ proceeds as follows: with $\U_0^k = \X(\W - \delta^{k-1}\Q^{k-1})$, we iterate for $i = 1,\dots, m$:
\begin{align}\label{eq:udpate qij}
   &  \U_i^k = \U_{i-1}^k - \x_{:,i} \otimes (\w_{i,:}- \delta^{k-1}\q_{i,:}^{k-1})
   \notag \\
 & \q^k_{i,:} = \text{clip}\left (  
       \left \lfloor \frac{(\U^k_i + \x_{:,i} \otimes \w_{i,:})^\top \x_{:,i} }{\delta^{k-1} \x_{:,i}^\top \x_{:,i}} \right\rceil, z, z+2^{b}-1
        \right)  \\
 &  \U_i^k = \U_i^k + \x_{:,i} \otimes (\w_{i,:}-\delta^{k-1} \q_{i,:}^k).  \notag
\end{align}
Here $\U_i^k$ maintains the quantization residuals across all columns.

\smallskip

\noindent \textbf{$\delta$-Update}. After obtaining the new bit-code matrix $\Q^{k}$, the scale factor can be updated by solving the following problem 
\[
    \min_\delta \|\delta \X\Q^k - \X \W \|^2.
\]
It is a standard convex quadratic optimization problem and has a closed-form solution 
\begin{equation}\label{eq:update delta}
    \delta^k = \frac{\left<\X\Q^{k}, \X\W\right>}{\|\X\Q^{k}\|^{2}}.
\end{equation}

 We summarize the COMQ algorithm for linear layers in Alg. \ref{alg:COMQ linear}. For a general neural network with multiple linear layers, we can apply the COMQ method in Alg. \ref{alg:COMQ linear} to each linear layer to obtain the quantized neural network.

\renewcommand{\algorithmicrequire}{\textbf{Input:}} 
\renewcommand{\algorithmicensure}{\textbf{Output:}}
\begin{algorithm}[H]
    \caption{COMQ for the per-layer quantization of one linear layer.}
    \label{alg:COMQ linear}
    \begin{algorithmic}[1] 
        \REQUIRE Pre-trained weights $\W \in \mathbb{R}^{m\times n}$, feature matrix $\X$, and iteration number $K$.

            \STATE \textbf{Initialize} $\delta^0$ and $\Q^0$
            \FOR{$k=1,\ldots,K$ }
                \FOR{$i=1,\ldots,m$}
                    \STATE{Update the coordinates $\{Q_{i,j}^k\}_{j\in[n]}$ or $\q^k_{i,:}$ as in \eqref{eq:udpate qij}}
                \ENDFOR
                \STATE Update the scaling factor $\delta^k$ as in \eqref{eq:update delta}
            \ENDFOR
            \STATE Let $\W_q = \delta^K \Q^K$
        \ENSURE Quantized weight $\W_q$.
    \end{algorithmic} 
\end{algorithm}

\subsection{Per-channel Quantization} 

Per-layer quantization uses the same scale factor for the whole weight matrix. However, the different columns of weight matrix may have very different value ranges, and this causes a large quantization error. For the per-channel quantization, each column has its own scale factor, which yields smaller quantization errors and less accuracy drop.

\begin{figure*}[htbp] 
\centering 
\includegraphics[height=0.45\textwidth]{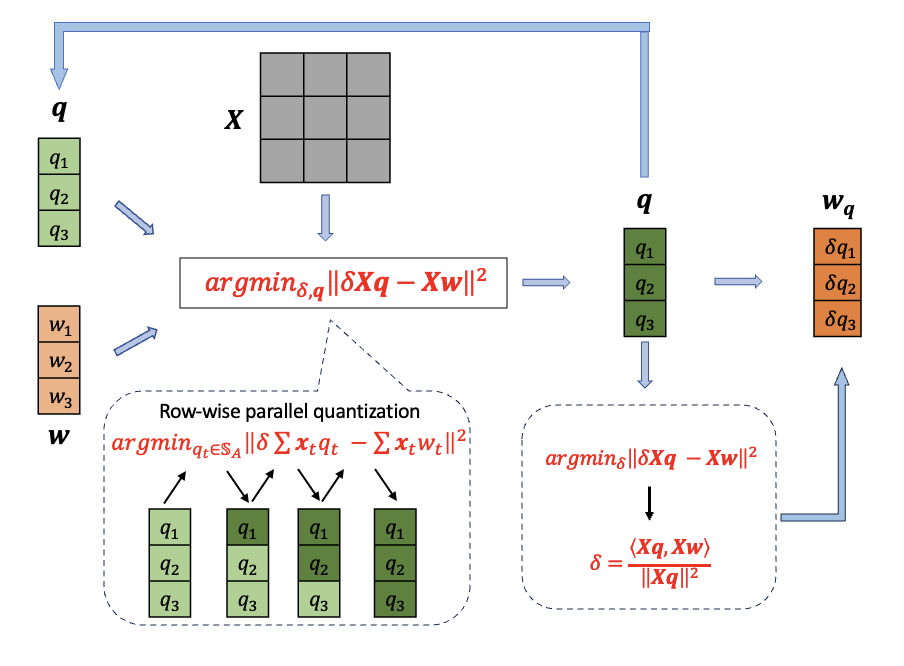} 
\caption{The workflow of COMQ for per-channel quantization.}
\label{fig:per-channel} 
\end{figure*}
\label{sc:channel}

The per-channel quantization aims to quantize the $j$-th column $\w_i$ of $\W$ to $\delta_j \q_j$, where $\q_j$ is bit-code vector.
For the weight matrix $\W$, the quantized weight $\W_q \in \mathbb{R}^{m\times n}$ in per-channel quantization format is 
\[
     \W_q = \Q \text{diag}(\boldsymbol{\delta})= \Q \text{diag}(\delta_1,\ldots,\delta_n)= (\delta_1 \q_1,\ldots,\delta_n \q_n).
\]
 Same as for per-layer quantization, the $j$-th column of $\W$ can be quantized independently by solving the following optimization problem:
 \begin{equation} \label{eq:minqi channel}
     \delta_j,\q_j = \arg \min_{\delta,\q} \|\delta \X\q - \X\w_j\|^2,\, j\in[n].
 \end{equation}
 

\smallskip


 \noindent \textbf{Initialization.} The scale factor is initialized as 
 $\delta_j^0=\lambda \frac{\max(\w_j)-\min(\w_j)}{2^{b}-1}$ for some $0<\lambda\le 1$. The $\lambda$ is to ensure we do not quantize most values to zero, especially for ultra-low bit quantization. The $\q_j^0$ is initialized as $\frac{\w_j}{\delta_j^0}$. 

\begin{figure*}[ht] 
\centering 
\includegraphics[width=1\textwidth]{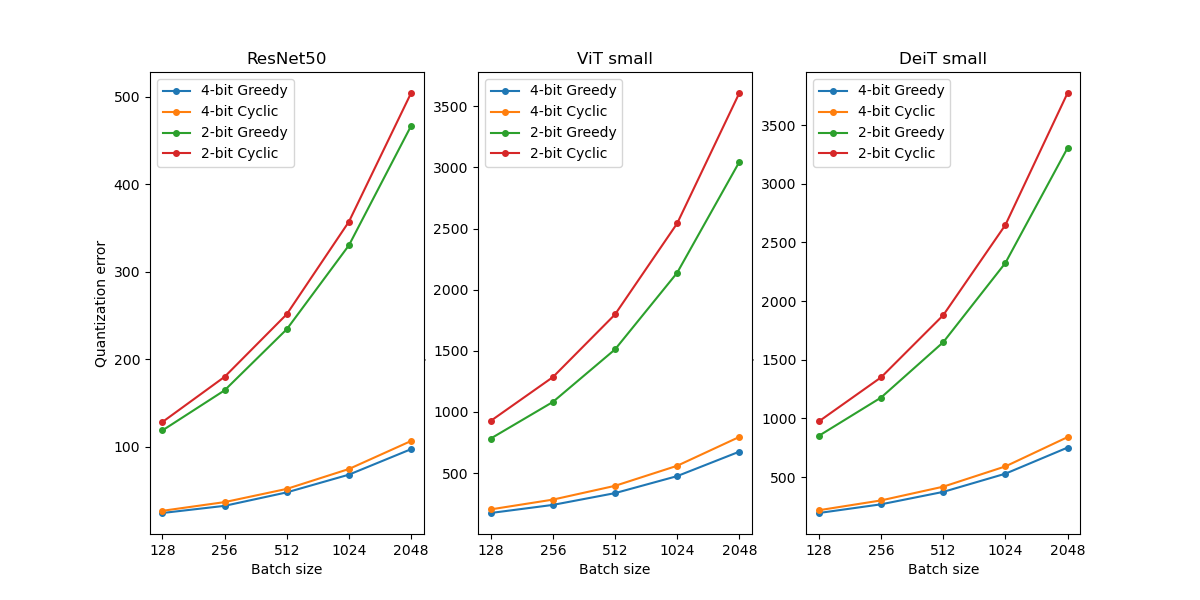} 
\caption{Comparisons of layer-wise quantization errors for cyclic and greedy orders.}
\label{fig:greedy_cyclic_compare} 
\end{figure*}

\smallskip


\noindent \textbf{$\Q$-Update}. Let $\boldsymbol{\delta}^{k-1}\in\R^n$ and $\Q^{k-1}\in\mathbb{S}^{m\times n}$ be the scaling factors and bit-code matrix produced by $(k-1)$-th iteration. Or equivalently, suppose $\W_q^{k-1} = \Q^{k-1} \text{diag}(\boldsymbol{\delta}^{k-1})$ is the current quantized weight matrix. 
The $\Q$-Update for per-channel quantization is substantially similar to the per-layer setting. The updated bit-code $Q_{i,j}^{k}$ is given by 
\begin{equation*} 
    Q^{k}_{i,j} = \text{clip} \left ( \left \lfloor \frac{\left< \x_{i}, \s^k_{i,j} \right>}{\delta_j^{k-1}\|\x_{i}\|^{2}} \right \rceil, z_j, z_j+2^{b}-1 \right ),
\end{equation*}
where $\s_{i,j}^k = \X \w_j - \delta_j^{k-1}(\sum_{t=1}^{i-1} Q_{t,j}^k \x_t + \sum_{t=i+1}^{m} Q_{t,j}^{k-1} \x_t)$, and $z_j$ is the zero-point for quantizing the $j$-th column. For the row-wise update of $\Q^k$, we simply follow \eqref{eq:udpate qij} with minor adaptions to the column-wise scaling. Using the same notations, we denote by $\w_{i,:}\in\R^n$ and $\q_{i,:}\in\R^n$ the $i$-th row of $\W$ and $\Q$, respectively, and denote by $\x_{:,i}\in\R^m$ the $i$-th column of $\X$. With $\U_0^k = \X(\W - \W_q^{k-1})$, we iterate for $i = 1,\dots, m$:
\begin{align}\label{eq:updateq channel}
   &  \U_i^k = \U_{i-1}^k - \x_{:,i} \otimes (\w_{i,:}- \boldsymbol{\delta}^{k-1} \odot \q_{i,:}^{k-1})
   \notag \\
 & \q^k_{i,:} = \text{clip}\left (  
       \left \lfloor \frac{(\U^k_i + \x_{:,i} \otimes \w_{i,:})^\top \x_{:,i} }{\x_{:,i}^\top \x_{:,i}} \oslash \boldsymbol{\delta}^{k-1} \right\rceil , \boldsymbol{z}, \boldsymbol{z} +2^{b}-1
        \right)  \\
 &  \U_i^k = \U_i^k + \x_{:,i} \otimes (\w_{i,:}-\boldsymbol{\delta}^{k-1} \odot \q_{i,:}^{k}),  \notag
\end{align}
where $\U_i^k$ maintains the quantization residuals and $\boldsymbol{z}$ is the vector of zero-points across all columns.

\smallskip
\noindent \textbf{$\boldsymbol{\delta}$-Update}. Having obtained $\Q^k$, we update the scaling factors as 

\begin{equation} \label{eq:updatedelta channel}
\delta^{k}_j = \frac{\left<\X\q_j^k, \X \w_j\right>}{\|\X\q_j^k\|^{2}}.
\end{equation}

To summarize, the workflow for per-channel quantization is depicted in Fig. \ref{fig:per-channel}, and Alg. \ref{alg:COMQ channel} describes COMQ for layer-wise PTQ with per-channel scaling. 
\vspace*{-10pt}
\begin{algorithm}[H]
    \caption{COMQ for the per-channel quantization of one linear layer}
    \label{alg:COMQ channel}
    \begin{algorithmic}[1] 
        \REQUIRE Pre-trained weights $\W \in \mathbb{R}^{m\times n}$, feature matirx $\X$, and iteration number $K$.
            \STATE \textbf{Initialize} $\boldsymbol{\delta}^0,\Q^0$ as described in Section \ref{sc:channel}
            \FOR{$k=1,\ldots,K$ }
                \FOR{$i=1,\ldots,m$}
                    \STATE{Update the coordinates $\{Q_{i,j}^k\}_{j\in[n]}$ or $\q^k_{i,:}$ as in \eqref{eq:updateq channel}}
                \ENDFOR
                \STATE Update the scaling factors $\boldsymbol{\delta}^k$ as in \eqref{eq:updatedelta channel}
            \ENDFOR
            \STATE Compute $\W_q = (\delta_1^K \q_1^K,\ldots,\delta_n^K \q_n^K)$
        \ENSURE Quantized weight $\W_q$.
    \end{algorithmic} 
\end{algorithm}

 \subsection{Greedy COMQ} 

Coordinate descent update the coordinates of a vector in the cyclic order (or index order) by default, which may not be optimal. To further reduce the quantization error and improve the performance of the quantized model, we propose a greedy update rule to determine the update order of the coordinates. The quantization error for a column with weight vector $\w$ can be written as 
 $\|\delta \X\q-\X\w\|^2 = \left\| \sum_{i=1}^m \delta q_{i} \x_i- w_{i} \x_i \right\|^2.$
The importance of quantization target $w_i \x_i$ can be heuristically measured by its magnitude. Hence, we choose to first update the coordinates with a larger magnitude. This order ensures that the most significant coordinates are first updated, and we have a sufficient quantization error decrease for each step. In practice, we sort $\|w_1 \x_1\|,\ldots,\|w_m \x_m\|$ in descending order to get the coordinate update order $i_1,\ldots,i_m$, permuting the index set $[m]$. Therefore, the greedy COMQ for quantizing the weight matrix $\W$ requires column-wise sorting of $(\boldsymbol{v}\odot |\w_1|,\dots,\boldsymbol{v}\odot|\w_n|)$, where $\boldsymbol{v} = (\|\x_1\|, \dots, \|\x_m\|)^\top$ is the column-wise Euclidean norm of $\X$ in descending order. Then, we permute the columns of $\W$ and rows of $\X$ individually according to the sorted indices, followed by the quantization process. After that, the inverse permutations are performed on $\W$ and $\X$ to restore the original index order.

To demonstrate the superiority of our proposed greedy order over the cyclic order, we compared the empirical layer-wise quantization errors $\|\X\W_q - \X\W\|$ across different architectures in Fig. \ref{fig:greedy_cyclic_compare}. Additionally, we also compared  their performance on ImageNet in Tab. \ref{table:cyclic} in the Appendix.



\begin{table}[ht]
\centering
\caption{Comparison of ImageNet Top-1 accuracy (\%) on ViTs using \emph{per-channel} weight-only uniform quantization.}
\label{table:wvit}
\begin{tabularx}{\textwidth}{l >{\centering\arraybackslash}X>{\centering\arraybackslash}X>{\centering\arraybackslash}X>{\centering\arraybackslash}X>{\centering\arraybackslash}X>{\centering\arraybackslash}X l} 
\toprule[1.5pt]
Method & WBit&ViT-S & ViT-B & DeiT-S & DeiT-B & Swin-T & Swin-S \\
\midrule
Baseline & 32 & 81.39 & 84.53 & 79.85 & 81.99 & 81.38 & 83.21\\
\midrule
FQ-ViT \cite{lin2021fq} & 4 & - & 81.09 & 76.23 & 79.92 & 78.81 & 81.89\\

PTQ4ViT \cite{yuan2021ptq4vit} & 4 & 72.34 & 72.06 & 77.50 & 80.07 & 78.46 & 80.24\\

Ours & 4 & $\textbf{80.35}$ & $\textbf{83.86}$ & $\textbf{78.98}$ & $\textbf{81.40}$ & $\textbf{80.89}$ & $\textbf{82.85}$\\
\midrule 
FQ-ViT \cite{lin2021fq} & 3 & - & 34.31 & 51.06 & 65.64 & 65.38 & 71.88\\

PTQ4ViT \cite{yuan2021ptq4vit} & 3 & 38.77 & 19.85 & 70.22 & 75.42 & 70.74 & 73.46\\

Ours & 3 & $\textbf{77.08}$ &  $\textbf{81.73}$ & $\textbf{77.47}$ & $\textbf{80.47}$ & $\textbf{79.31}$ & $\textbf{81.95}$\\
\midrule

Ours & 2 & $\textbf{52.44}$ &  $\textbf{65.69}$ & $\textbf{67.19}$ & $\textbf{77.14}$ &$\textbf{74.05}$ & $\textbf{78.02}$\\
\bottomrule[1.5pt]
\end{tabularx}

\end{table}
\vspace*{-20pt}
\begin{table}[h]
\centering
\caption{Comparison of ImageNet Top-1 accuracy (\%) on ViTs using \emph{per-channel} full uniform quantization.}
\label{table:awvit}
\begin{tabularx}{\textwidth}{l>{\centering\arraybackslash}X>{\centering\arraybackslash}X>{\centering\arraybackslash}X>{\centering\arraybackslash}X>{\centering\arraybackslash}X>{\centering\arraybackslash}Xl} 
\toprule[1.5pt]
Method & Bit (W/A) &ViT-S & ViT-B & DeiT-S & DeiT-B & Swin-S \\
\midrule[1.2pt]
BRECQ \cite{li2021brecq} & 4/4 & 12.36 & 9.68 & 63.73 & 72.31 & 72.74\\

QDrop \cite{wei2022qdrop} & 4/4 & 21.42 & 47.30 & 68.27 & 72.60 & 79.58\\

APQ-ViT \cite{ding2022towards} & 4/4 & 47.95 & 41.41 & 43.55 & 67.48 & 77.15\\

RepQ-ViT \cite{li2023repq} & 4/4 & 65.05 & 68.48 & 69.03 & 75.61 & 79.45\\
Ours & 4/4 & $\textbf{71.47}$ & $\textbf{78.27}$ & $\textbf{72.17}$ & $\textbf{78.72}$ &$\textbf{81.19}$\\
\midrule 

Ours & 2/4 & $\textbf{30.11}$ &$\textbf{45.33}$&$\textbf{53.20}$& $\textbf{71.90}$ & $\textbf{75.37}$\\
\bottomrule[1.5pt]
\end{tabularx}

\end{table}
\vspace*{-20pt}
\section{Experiments}
In this part, we demonstrate the superiority of our method on various neural architectures, including ResNets \cite{he2016deep}, MobileNetV2 \cite{howard2017mobilenets}, and ViTs \cite{dosovitskiy2020image}. Sec. \ref{sc:implementation} provides the implementation details. In Sec. \ref{sc:compare}, we compare the proposed COMQ method with the state-of-the-art PTQ methods. In Sec. \ref{sc:ablation}, we conducted extensive ablation studies to comprehensively analyze the properties of our algorithm, including the influence of various batch sizes, the number of iterations and the runtime of different batch sizes.

\subsection{Implementation Details}
\label{sc:implementation}
We validate the performance of COMQ on the ImageNet \cite{imagenet_cvpr09} dataset. Evaluation metrics are the precision of the top-1 and top-5 quantized models in the validation data set. With the pre-trained float models from PyTorch and BRECQ \cite{li2021brecq}, we quantize various CNNs on ImageNet including ResNet18\cite{he2016deep}, ResNet50\cite{he2016deep}, MobileNetV2\cite{howard2017mobilenets} and Vision Transformers including ViT\cite{dosovitskiy2020image}, DeiT\cite{touvron2021training} and Swin\cite{liu2021swin}. We used batch size 2048 for ResNet18, ResNet50, DeiT, and batch size 1024 for MobileNetV2, ViTs, Swin. We ran our experiments on an Nvidia RTX 3090 GPU with 24G GPU memory for convolutional neural nets and on an Nvidia A40 GPU with 48G GPU memory for ViTs. 

\subsection{Comparison with the State-of-the-arts}
\label{sc:compare}
\subsubsection{Vision Transformer}
We evaluated COMQ with per-channel quantization in Alg. \ref{alg:COMQ channel} on various architectures of vision transformers, such as ViT, DeiT, and Swin, and compared its performance with previous methods PTQ4ViT \cite{frantar2022gptq} and FQ-ViT \cite{lin2021fq}. These methods exhibit good performance in $4$-bit quantization but perform poorly in lower $3$-bit quantization and do not work for $2$-bit quantization. Tab. \ref{table:wvit} compares our method with these methods for weight quantization. Tab. \ref{table:awvit} further quantizes the activations into $4$-bit. Across all ViT models and precision levels, COMQ has better performance than the existing PTQ methods. In particular, we achieve a remarkable high accuracy for $3$-bit quantization. To be noted, our method is the first to push the precision down to $2$-bit (2W32A and 2W4A) quantization while maintaining high accuracy. Note that for the activation quantization, we adopted the reparameterization method proposed in \cite{li2023repq} and incorporated it into COMQ.

\setlength{\tabcolsep}{9pt}
\begin{table}[h!]
\centering
\caption{Comparison of ImageNet Top-1 accuracy (\%) on ResNets and MobileNetV2 with \emph{per-layer} weight-only uniform quantization.}
\label{table:4}
\begin{tabular}{lcccc} 
\toprule[1.5pt]
Method & WBit& ResNet18 & ResNet50 & MobileNetV2  \\
\midrule
Baseline & 32 & 69.76 & 76.13 & 72.49  \\
\midrule
AdaRound \cite{nagel2020up} & \multirow{4}{*}{4} & 66.56& -& - \\

Bit-spilt \cite{wang2020towards} &  & 68.31 & 70.56 & - \\

AdaQuant \cite{hubara2020improving} &  & 68.12 &  74.68 & 44.78 \\

Ours$^\dagger$ & & 68.76 &  75.19 & 70.51\\
Ours & & $\textbf{69.26}$ &  $\textbf{75.50}$ & $\textbf{70.49}$\\
\midrule
Bit-spilt \cite{wang2020towards} & \multirow{3}{*}{3} & 64.77 &  66.98 & -\\

AdaQuant \cite{hubara2020improving} &  & 59.21 &  64.98 & 12.56  \\

Ours$^\dagger$ & & 65.63 & 68.23 & 62.36 \\

Ours & & $\textbf{65.72}$ &  $\textbf{71.64}$ & $\textbf{62.43}$\\


\bottomrule[1.5pt]
\end{tabular}

\end{table}
\vspace*{-15pt}
\subsubsection{Convolutional Neural Networks}

Beyond Vision Transformer, we also evaluated COMQ with per-layer and per-channel quantization on CNN models and compared it with state-of-the-art uniform post-training quantization methods, including Bit-split \cite{wang2020towards}, AdaRound \cite{nagel2020up}, AdaQuant \cite{hubara2021accurate}, BRECQ \cite{li2021brecq} and OBQ \cite{frantar2022optimal}.

\smallskip

\noindent $\textbf{Per-layer Quantization}$.
Per-layer quantization is more computationally efficient but per-layer quantization in existing methods typically leads to high performance degradation. In contrast, our greedy COMQ for per-layer quantization in Alg. \ref{alg:COMQ linear} yields promising results, particularly on ResNets. With the float models from PyTorch, we compare COMQ with state-of-the-art uniform per-layer PTQ methods, such as AdaQuant \cite{nagel2020up} and Bit-split \cite{wang2020towards}. The results are presented in Table \ref{table:4}. For $4$-bit, COMQ achieves a mere 1.0\% accuracy drop in ResNet18 and a 0.94\% accuracy drop in ResNet50, outperforming other methods for per-layer PTQ. More remarkably, for $3$-bit quantization, COMQ achieves a 4.13\% and an 7.9\% accuracy drop on ResNet18 and ResNet50, respectively, surpassing all competing methods. The cyclic COMQ for per-layer quantization is highlighted by a $^\dagger$ symbol. 

\renewcommand{\arraystretch}{0.75}
\setlength{\tabcolsep}{10pt}
\begin{table}[h]
\centering
\caption{Comparison of ImageNet Top-1 accuracy (\%) on ResNets using \emph{per-channel} weight-only uniform quantization.}
\label{table:3}
\begin{tabular}{lcccc} 
\toprule[1.5pt]
Method &Bit (W/A)&ResNet18 & ResNet50  \\
\midrule
Baseline & 32/32 & 71.00 & 76.63 \\
\midrule

Bit-split \cite{wang2020towards} & 4/32 & 69.11 & 75.58  \\

AdaRound \cite{nagel2020up} & 4/32 & 68.71 & 75.23  \\ 

FlexRound \cite{lee2023flexround} & 4/32 & 70.28 &  75.95  \\

BRECQ \cite{li2021brecq} & 4/32 & 70.70 &  76.29  \\


OBQ \cite{frantar2022optimal} & 4/32 & 70.42 & 76.09  \\

Ours & 4/32 & $\textbf{70.83}$ &  $\textbf{76.38}$  \\
\midrule
Bit-split \cite{wang2020towards} & 3/32 & 66.75& 73.24  \\

AdaRound \cite{nagel2020up} & 3/32 & 68.07 & 73.42  \\ 

FlexRound \cite{nagel2020up} & 3/32 & 68.65 & 74.38  \\ 

BRECQ \cite{li2021brecq} & 3/32 & \textbf{69.81} &  75.61  \\

OBQ \cite{frantar2022optimal} & 3/32 & 68.96 & 74.23  \\

Ours & 3/32 & 69.63 &  $\textbf{75.73}$  \\
\midrule
AdaRound \cite{nagel2020up} & 2/32 & 55.96 & 47.95  \\ 

FlexRound \cite{nagel2020up} & 2/32 & 62.57 & 63.67  \\

BRECQ \cite{li2021brecq} & 2/32 & $\textbf{66.30}$ &  $\textbf{72.40}$  \\

OBQ \cite{frantar2022optimal} & 2/32 & 63.15 & 68.49 \\

Ours & 2/32 & 64.52 &  70.32  \\
\bottomrule[1.5pt]
\end{tabular}

\end{table}

\noindent $\textbf{Per-channel Quantization}.$ 
We evaluated greedy COMQ for per-channel quantization described in Alg. \ref{alg:COMQ channel} and compared its performance with state-of-the-art PTQ methods with uniform quantization, such as Bit-split \cite{wang2020towards}, AdaRound \cite{nagel2020up}, AdaQuant \cite{hubara2021accurate}, BRECQ \cite{li2021brecq}, and OBQ \cite{frantar2022optimal}. The results are shown in Tab. \ref{table:3} and Tab. \ref{table:aw}. For $4$-bit quantization, our method almost attains lossless accuracy and outperforms the existing methods on both ResNet18 and ResNet50. For $3$-bit, Greedy COMQ exhibits a comparable accuracy on ResNet18 and slightly outperforms existing methods on ResNet50. Even for $2$-bit quantization, our method exhibits promising results on both models. Although the results are inferior to those of BRECQ, we remark that the implementations of BRECQ and other similar algorithms rely on costly back-propagation.
For instance, the runtime of COMQ for quantizing ResNet18 on an Nvidia RTX 3090 GPU is merely 20 seconds, in stark contrast to the nearly 50 minutes required by BRECQ.
\setlength{\tabcolsep}{11pt}
\begin{table}[ht]
\centering
\caption{Comparison of ImageNet Top-1 accuracy (\%) on ResNets using \emph{per-channel} full quantization.}
\label{table:aw}
\begin{tabular}{l|c|ccc} 
\toprule[1.5pt]
Method &Bit (W/A)&ResNet18 & ResNet50  \\
\midrule
Baseline & 32/32 & 71.00 & 76.63 \\
\midrule

Bit-split \cite{wang2020towards} & 4/4 & 67.56 & 73.71  \\

AdaRound \cite{nagel2020up} & 4/4 & 69.20 & 72.79  \\ 

FlexRound \cite{lee2023flexround} & 4/4 & 69.26 & 75.08  \\

BRECQ \cite{li2021brecq} & 4/4 & 69.60 &  75.05  \\

QDrop \cite{wei2022qdrop} & 4/4 & 69.62 & 75.45  \\


Ours & 4/4 & $\textbf{69.70}$ &  $\textbf{75.46}$  \\

\bottomrule[1.5pt]
\end{tabular}

\end{table}

\begin{table}[ht]
\centering
\caption{Accuracy vs Batchsize for 4W32A per-channel PTQ.}
\label{table:bs}
\resizebox{\linewidth}{!}{
\begin{tabular}{lcccccc} 
\toprule[1.2pt]
Batch size& 128 &256 & 512 & 1024 & 2048  & FP Baseline\\
\midrule
ResNet18 & 69.34 & 69.28 & 69.47 & 69.54 & 69.75 & 69.76 \\
\midrule
ResNet50 & 76.01 & 76.05 & 76.04 &76.08& 76.09 & 76.13\\
\midrule
ViT-B & 83.53 & 83.51 & 83.73 &83.86& 83.79 & 84.53\\
\bottomrule[1.2pt]
\end{tabular}
}

\end{table}
\begin{table}[ht]
\centering
\caption{Accuracy vs Iteration Number $K$ for 4W32A per-layer PTQ.}
\label{table:iter}
\resizebox{\linewidth}{!}{
\begin{tabular}{lcccccc} 
\toprule[1.2pt]
K  & 1 & 2 & 3 & 4 & 5 & FP Baseline\\
\midrule
ResNet18 & 68.52 & 68.65 & 68.76 & 68.65 & 68.57 & 69.76\\
\midrule
ResNet50  & 74.87 & 75.08 & 75.19 &75.16& 75.15 & 76.13\\
\bottomrule[1.2pt]
\end{tabular}
}

\end{table}

\subsection{Ablation Study}
\label{sc:ablation}
In this subsection, we examine how various factors influence the performance and efficacy of our algorithm. We evaluate our method on CNNs with float models from PyTorch and Vit-B with float model from open source Timm.

\smallskip

\noindent $\textbf{Batch size}.$
All experiments are for per-channel PTQ. The number of operations (dot product and rounding) required by COMQ depends on the number of weights, not on the batch size. The larger batch size will result only in dot products performed in a higher dimension, which can still be efficient. We see that COMQ also performs reasonably well with a small batch size as shown by Tab  \ref{table:bs}. 

\smallskip

\noindent $\textbf{Iteration number K}.$
All experiments are for per-layer PTQ. As shown in Table \ref{table:iter}, more iterations do not necessarily equate to better results. Typically, the optimal solution is achieved when $K=3$ or $4$, additional iterations may not deliver significant improvements.

\section{Concluding Remarks}
In conclusion, our research presented COMQ, a novel coordinate-wise minimization algorithm designed for the post-training quantization (PTQ) of convolutional neural nets and transformers. COMQ solves the minimization of the layer-wise squared reconstruction error, treating all quantization parameters within the same layer, including weights and floating-point scalars, as variables in the error function.
One notable feature of COMQ is its efficiency in each iteration, which involves only dot products and rounding operations. This simplicity distinguishes COMQ from existing PTQ approaches, making it a low-cost alternative. Importantly, the algorithm requires no hyper-parameter tuning to achieve state-of-the-art performance in image classification tasks.
Our experiments demonstrate that COMQ surpasses existing methods, especially in the ultra-low bit-width regime, showcasing superior uniform PTQ results on ImageNet for Vision Transformers. This highlights the effectiveness of COMQ in achieving optimal quantization outcomes with minimal computational overhead, thus contributing to the advancement of PTQ techniques for DNNs. Future work includes the extension of PTQ based on prediction difference metric \cite{liu2023pd} and for multimodal models such as vision-language models \cite{wortsman2024stable}. It is also possible to combine per-layer and per-channel quantization strategies into a mix-precision quantization framework \cite{chauhan2023post}.

\section*{Acknowledgement}
PY and AZ were partially supported by NSF grants DMS-2208126 and IIS-2110546. ZY was supported by NSF grant DMS-2110836 and a start-up grant from SUNY Albany. JX was supported by NSF grants DMS-2151225 and DMS-2219904, and a Qualcomm faculty gift award. XL was supported by NSF grant CCSS-2348046.

\clearpage  

%
%
\bibliographystyle{splncs04}
\bibliography{main}
\newpage
\section*{Appendix: More Ablation Study}

We compare greedy update order and cyclic update order for the per-channel quantization in Algorithm \ref{alg:COMQ channel}. We evaluate the two update orders on five widely used models: ResNet18, ResNet50, ViT-S, DeiT-S, and Swin-T. The pre-trained ResNet family models are sourced from the PyTorch platform, and the pre-trained vision transformer models are trained on ImageNet using the open-source Timm library. The results are shown in Table \ref{table:cyclic}. We can find that the greedy update order outperforms the cyclic update order in all test cases across all models and precisions, demonstrating that the proposed greedy update order greatly improves the performance of quantized models. Furthermore, the performance improvement is more significant for larger models at lower bit-widths.
\vspace{-20pt}
\begin{table}[ht]
\centering
\caption{ImageNet results for cyclic and greedy COMQ with weight quantization.}
\label{table:cyclic}
\setlength{\tabcolsep}{2.5pt}
\begin{tabular}{c|c|ccccc} 
\toprule[1.5pt]
Method & Bits & RN18 & RN50 & ViT-S & DeiT-S & Swin-T \\
\midrule
FP & 32 & 71.00 & 76.63 & 81.39 & 81.99 & 81.38 \\
\midrule
Cyclic & \multirow{2}{*}{4} & 70.71& 76.29 & 80.16 & 78.94 & 80.85\\

Greedy & &$\textbf{70.83}$ & $\textbf{76.38}$ & $\textbf{80.35}$ & $\textbf{78.98}$ & $\textbf{80.89}$\\
\midrule 
Cyclic & \multirow{2}{*}{3} &69.53 & 75.58 & 76.58 & 77.20 & 78.81\\
Greedy & & $\textbf{69.63}$ &  $\textbf{75.73}$ & $\textbf{77.08}$ & $\textbf{77.47}$ & $\textbf{79.31}$\\
\midrule
Cyclic & \multirow{2}{*}{2} & 64.24 & 69.21 & 49.27 & 65.46 & 73.05\\
Greedy & & $\textbf{64.52}$ &  \textbf{70.32} & $\textbf{52.44}$&\textbf{67.19} & $\textbf{74.05}$\\
\bottomrule[1.5pt]
\end{tabular}
\end{table}
\vspace{-10pt}

The following Table \ref{table:runtime} shows the runtime comparison on ResNet50 for the vector-wise update.
\vspace{-10pt}
\begin{table}[ht]
\centering
\caption{Runtime of 4W32A COMQ (greedy), which is faster than the gradient-based methods.}
\vspace{-10pt}
\label{table:runtime}
\renewcommand{\arraystretch}{1.0}
\begin{tabular}{lcccc} 
\toprule[1.5pt]
Model & AdaRound & BRECQ & OBQ & COMQ    \\
\midrule
ResNet50 & 55 min & 53 min & 65 min & 12 min \\
\bottomrule[1.5pt]
\end{tabular}

\end{table}
\vspace{-10pt}

Table \ref{table:3} shows the performance of our algorithm on Swin ViTs for different values of $\lambda$ in 2 bits. It is clear that the results of $lambda=0.71$ (near-optimal $\lambda$) are much better than that of $\lambda=1$.
\vspace{-10pt}
\begin{table}[ht]
\centering
\caption{ImageNet accuracy for different $\lambda$ initialization. $\lambda=0.71$ is empirically (near-)optimal for Swin ViTs.}
\label{table:3}
\renewcommand{\arraystretch}{0.8}
\begin{tabular}{c|c|ccccc} 
\toprule[1.5pt]
$\lambda$ & Bits  & Swin-T & Swin-S \\
\midrule
1 & \multirow{2}{*}{2} & 65.05& 70.06 \\
0.71 & &$\textbf{74.05}$ & $\textbf{78.02}$ \\
\midrule
\midrule
FP & 32 & 81.38  & 83.21 \\
\bottomrule[1.5pt]
\end{tabular}

\end{table}

\end{document}